\documentclass{article}

\PassOptionsToPackage{numbers, compress}{natbib}

\usepackage[preprint]{neurips_2025}

\usepackage{graphicx}
\usepackage[utf8]{inputenc} % allow utf-8 input
\usepackage[T1]{fontenc}    % use 8-bit T1 fonts
\usepackage{hyperref}       % hyperlinks
\usepackage{url}            % simple URL typesetting
\usepackage{booktabs}       % professional-quality tables
\usepackage{amsfonts}       % blackboard math symbols
\usepackage{nicefrac}       % compact symbols for 1/2, etc.
\usepackage{microtype}      % microtypography
\usepackage{xcolor}         % colors
\usepackage{caption}
\usepackage{algorithm}
\usepackage{algorithmic}
\usepackage{amsmath}
\usepackage{cleveref}

\title{DiffDecompose: Layer-Wise Decomposition of Alpha-Composited Images via Diffusion Transformers}

\author{
Zitong Wang$^{1}$\footnotemark[1]  \quad
Hang Zhao$^{1}$\thanks{Equal contribution.}  \quad
Qianyu Zhou$^{1}$\footnotemark[2] \quad
Xuequan Lu$^{2}$ \quad
Xiangtai Li$^{3}$ \quad
Yiren Song$^{4}$\thanks{Corresponding author.} \\
$^1$  Jilin University, China \hspace{0.2cm}
$^2$  University of Western Australia, Australia \\
$^3$  Nanyang Technological University, Singapore \hspace{0.2cm}
$^4$  National University of Singapore, Singapore\hspace{0.2cm}
}

\begin{document}

\maketitle
\vspace{-8mm}

\begin{figure}[ht]
    \centering
    \includegraphics[width=1.0\textwidth]{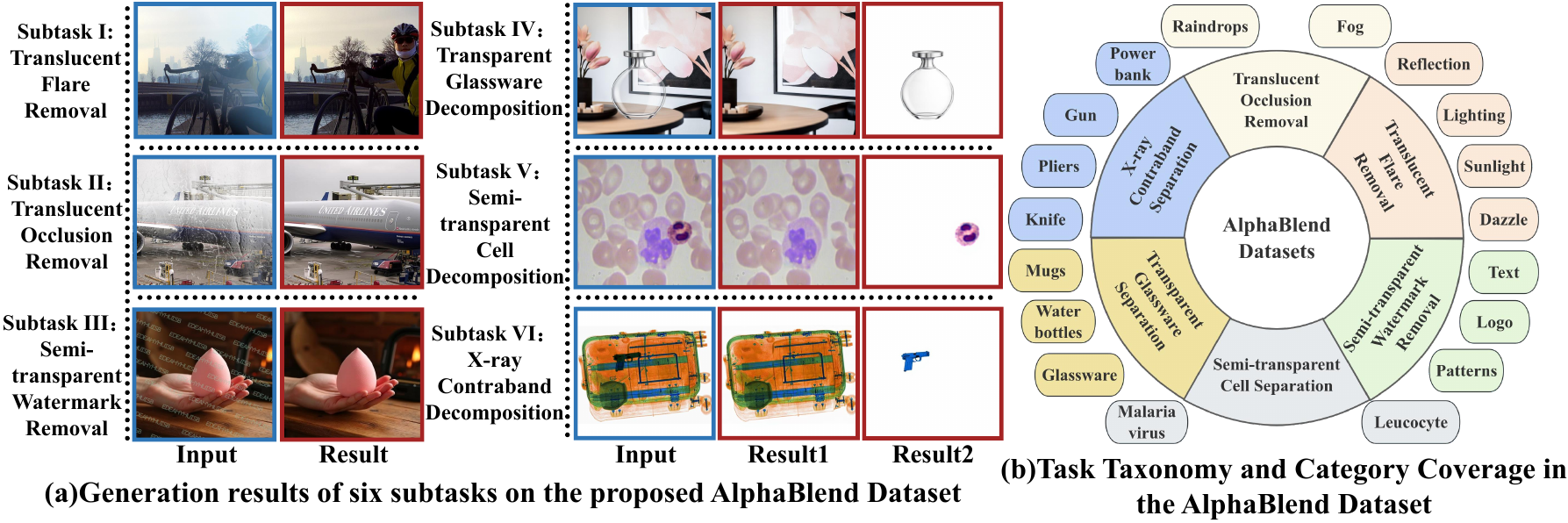}
    \caption{
    We propose a novel generative task, Layer-Wise Decomposition of Alpha-Composited Images, to recover constituent layers from single overlapped images under the condition of semi-transparent or transparent layer non-linear occlusion. 
    We introduce the AlphaBlend dataset, the first large-scale dataset for transparent and semi-transparent layer decomposition to support six real-world subtasks. (a) shows generation results on alpha layer removal (I–II), semi-transparent and transparent layer separation (III–IV), and complex non-linear alpha-blend decomposition (V–VI).  (b) highlights the dataset's broad coverage across categories \emph{e.g.,} flare, fog, glassware, X-ray contraband. }
    \label{fig1}
\end{figure}

\begin{abstract}

Diffusion models have recently achieved great success in various generation tasks, such as object removal. 
Nevertheless, existing image decomposition methods struggle to disentangle semi-transparent or transparent layer occlusions due to mask prior dependencies, static object assumptions, and the lack of datasets. 
In this paper, we explore a novel task: \emph{Layer-Wise Decomposition of Alpha-Composited Images}, aiming to recover constituent layers from single overlapped images under the condition of semi-transparent/transparent alpha layer non-linear occlusion. 
To address challenges in layer ambiguity, generalization, and data scarcity, we first introduce \emph{AlphaBlend}, the first large‑scale and high-quality dataset for transparent and semi‑transparent layer decomposition, supporting six real‑world subtasks (\emph{e.g.,} translucent flare removal, semi-transparent cell decomposition, glassware decomposition). 
Building on this dataset, we present \emph{DiffDecompose}, a diffusion transformer-based framework that learns the posterior over possible layer decompositions conditioned on the input image, semantic prompts, and blending type. 
Rather than regressing alpha mattes directly, DiffDecompose performs In‑Context Decomposition, enabling the model to predict one or multiple layers without per‑layer supervision, and introduces Layer Position Encoding Cloning to maintain pixel‑level correspondence across layers. 
Extensive experiments on the proposed AlphaBlend dataset and public LOGO dataset verify the effectiveness of DiffDecompose. The code and dataset will be available upon paper acceptance. 
Our code will be available at: \url{https://github.com/Wangzt1121/DiffDecompose}.

\end{abstract}

\section{Introduction}

With the development of diffusion models, image-to-image \cite{richardson2021encoding, ding2024enhance, xu2024ufogen, yang2025unified} and text-to-image \cite{betker2023improving, esser2024scaling, chen2023pixart} generation methods have been extensively studied and have a wide applications in image inpainting \cite{chen2024zero, sun2023alphaclip, xie2022smartbrush, wu2024towards}, video generation \cite{bar2024lumiere, wu2023tune, zhou2022magicvideo, wu2024motionbooth}, and image segmentation \cite{minaee2021image, amit2021segdiff, gu2024diffusioninst} to obtain high-quality images and user-friendly interfaces.
Most existing approaches rely on mask-based constraints to guide the model toward modifying or preserving specific image regions, or assume temporal consistency by treating moving object information as static to infer future or missing frames. Despite their success, these methods face two major limitations when applied to more complex and realistic scenarios. Firstly, the assumption of known separation regions or static objects is too simple, making it difficult to apply to more complex scenarios. Secondly, the noise in the diffusion model's forward phase may destroy the image's original information, reducing the image context's usability.

Recently, a new type of method, termed layer-wise decomposition \cite{tudosiu2024mulan, yang2024generative, zhang2024transparent, lee2024generative, chen2025transanimate,bai2025layer, gu2023factormatte}, has gradually entered the field of computer vision and become a potential method to solve the problem of these generative models. The core objective of layer-wise decomposition is to decompose a single composited image into its constituent layers, each containing the foreground object, its associated alpha matte (transparency), and potential depth ordering. Such a decomposition enables more granular control over individual layers, allowing for other tasks such as image inpainting \cite{shi2024foodfusion}, matting \cite{lu2025efficient}, scene understanding \cite{yang2024layerpano3d}, and content creation \cite{xie2025anywhere}. 
Nonetheless, these decomposing methods will fail to recover the accurate background when the foreground itself is layer-level semi-transparent or transparent, since they still rely on region-based assumptions.
Most importantly, current methods, such as image inpainting \cite{yu2023inpaint, sdxl, ekin2024clip, jiang2025smarteraserremoveimagesusing} and layer decomposition \cite{zhang2024transparent, yang2024generative}, cannot generalize to the real world to tackle arbitrary alpha-blended images with diverse contents and complex layer structures.

In this paper, our goal is to delve into a novel task that focuses on \emph{semi-transparent or transparent layer-wise decomposition}.
As shown in Figure \ref{fig1} (a), this novel task does not require any mask-based information for background assumption and no clear visual contrast to provide depth information, and under such conditions, a natural question is \textbf{how to decompose these transparent or semi-transparent layers from the background ?} Naturally, a single composited image often corresponds to multiple plausible decompositions due to the complex entanglement of color and transparency across layers, leading to additional difficulties in reconstructing the original scene. Therefore, this proposed task faced three main challenges not addressed in prior works: (1) \emph{Layer ambiguity and coupling of color and transparency}: foreground and background occupy the same visual plane and lack the separability via depth or edge contrast, thereby introducing significant layer ambiguity. Furthermore, critical foreground details may be partially embedded in the alpha channel, rendering them inaccessible through RGB information alone. (2) \emph{Generalization in real-world scenarios}: The above methods directly remove the pixel values of a specific area and use the network to generate new pixel values based on background reasoning. Such a manner largely overlooks the exploration of pixel-level correlations between the occluded area and the background, leading to the failure of generalization in various realistic scenarios. (3) \emph{Lack of large-scale dataset}: A significant challenge in layer decomposition is the lack of large-scale and high-quality datasets. Current 
datasets \cite{tudosiu2024mulan, yang2024generative} are often generated from AI, and the final overlapped images tend to have a pixel-level difference from the initial foreground and background, resulting in the failure of layer-wise decomposition.

Motivated by the above fact, we propose the first large-scale, high-quality dataset, namely \emph{AlphaBlend}, specifically designed for the decomposition of semi-transparent and transparent layers, where a class of scenarios frequently encountered in real-world applications but not previously presented in existing benchmarks. As shown in Figure \ref{fig1} (b), AlphaBlend has two characteristics: Firstly, it is the first dataset to support six different semi-transparent/transparent decomposition subtasks, and comprehensively constructed in modeling nonlinear compositional behavior, layer entanglement, and alpha-driven visual ambiguity. Secondly,  this dataset contains various styles and content, such as security, glass replacement, cell separation, \emph{etc.}, and each domain contains around 5000-10000 training images and 300-500 test images, which is closer to the real-world scenarios. The dataset will be released to the public to support further exploration of this new task upon paper acceptance.

Based on on the novel task and the AlphaBlend dataset, we propose a diffusion Transformer-based framework \emph{DiffDecompose} to probabilistically infer multiple plausible layer decompositions conditioned on context, which has two novel points: (1) Rather than explicitly regressing alpha mattes, we reformulate the task as learning the posterior distribution over compositional layers, conditioned on the observed image, semantic prompts, and blending type. (2) We design the In-Context Decomposition to enable the model to effectively predict single-layer or multi-layer results under given alpha-composited image conditions across various alpha compositional contexts without explicit supervision for each individual layer, and devise Layer Position Encoding Cloning (LPEC) to preserve the pixel-level correspondence. Our main contributions are three-fold:

\begin{itemize} 
\item We present a novel generative task, namely Layer-Wise Decomposition of Alpha-Composited Images for the semi-transparent/transparent scenarios, and reformulate this task from a novel perspective as a probabilistic problem of generating plausible layer decompositions. 
\item We introduce the AlphaBlend dataset, the first large-scale, high-quality dataset that supports six semi-transparent/transparent layer-wise decomposition subtasks, \emph{i.e.,} translucent flare removal, translucent occlusion removal, semi-translucent watermark removal, transparent glassware decomposition, semi-transparent cell decomposition, X-ray contraband decomposition. 
\item We propose a diffusion Transformer-based framework, namely DiffDecompose, and we are the first, to our best knowledge, to leverage the DiT for in-context decomposition without any implicit supervision. Concretely, we devise the Layer Position Encoding Cloning to realize the pixel-level alignment between different layers. 
Extensive experiments on the proposed AlphaBlend dataset and public LOGO dataset verify the effectiveness of DiffDecompose. 

\end{itemize}

\section{Related Work}
\vspace{-0.1cm}
\textbf{Diffusion Model.}
Diffusion Models (DMs) \cite{song2020denoising, rombach2021highresolution} have significantly advanced image generation and editing, surpassing GANs \cite{chen2019improved, liu2018image, pathak2016context, liu2018image, liu2022partial, wang2024semflow} in tasks such as artistic synthesis \cite{chen2024inverse}, instruction-guided transformation \cite{brooks2023instructpix2pix}, and controllable editing \cite{hui2024, yu2025}. Recent extensions have explored applications in 3D modeling \cite{li2025garment} and scene-consistent generation \cite{pan2025model}, showcasing the broad generative capacity of diffusion models. Despite these advances, diffusion models remain an open and ill-posed challenge, particularly in scenarios involving transparency, semi-transparency, and nonlinear blending. In this work, we present a novel generative task, namely \emph{Layer-Wise Decomposition of Alpha-Composited Images}, and propose a diffusion-based framework \emph{DiffDecompose}, to decompose complex semi-transparent/transparent layer-level composited images into plausible foreground and background layers. To support this novel task, we construct a first, large-scale and high-quality dataset \emph{AlphaBlend}, which mainly contains layer-level semi-transparent or transparent images.

\textbf{Controllable Generation.}
Recent advances in controllable diffusion models \cite{zhang2023adding, ye2023ip-adapter, sdxl, peebles2023, wu2024motionbooth} have enabled powerful image editing capabilities, particularly in inpainting. Existing approaches are typically categorized into text-based and mask-based methods. Text-based methods \cite{yu2024promptfix, zhuang2024task, xie2022smartbrush} rely on prompt engineering to achieve multi-task unification, but their performance is highly sensitive to prompt accuracy, often resulting in degraded outputs when instructions are imprecise \cite{xu2025pixel}. Mask-based methods \cite{yu2023inpaint, sdxl, ekin2024clip} focus on preserving background integrity via accurate region masks, yet they introduce noise to masked areas and perform iterative denoising, which weakens the conditioning signal and limits performance in complex blending scenarios. 
To this end, we extend a noise-free conditioning paradigm inspired by OminiControl \cite{tan2025}, from a novel point to reformulate the object occlusion task \cite{yu2024promptfix, ekin2024clip, sun2023alphaclip} or layer decomposition task \cite{tudosiu2024mulan, yang2024generative} as a probabilistic generation task, where the goal is to model the posterior distribution over plausible layer decompositions conditioned on the observed composite image. Under this definition, not only can we solve the common object-level or channel-level occlusion task, but also extend this idea to our proposed layer-level semi-transparent/transparent task.

\textbf{In-context Learning.}
In-context learning, originally developed in large language models \cite{mann2020language, zhang2025incontextedit}, has recently been extended to vision tasks such as object editing and region manipulation \cite{zhang2025incontextedit, wang2025explore, song2025insert, bai2024sequential, zhang2023makes, chen2025edittransfer, song2025makeanything, fang2023explore}. While prior works leverage contextual cues for object insertion or style transfer, none have addressed layer-wise decomposition, particularly in semi-transparent or transparent settings. 
In this paper, we introduce In-Context Decomposition (ICD), which enables the model to infer multiple compositional layers based on contextual prompts without explicit supervision. 
Furthermore, we propose a Layer Position Encoding Cloning (LPEC) strategy to preserve spatial alignment and reinforce information consistency between the decomposed layers and the original image.

\vspace{-0.1cm}
\section{Dataset and Methodology}
\vspace{-0.1cm}
\subsection{AlphaBlend Dataset Construction}\label{sec3}
To support the study of image compositing under complex transparency conditions, we introduced the AlphaBlend Dataset, which contains six sub-datasets for six different subtasks, including: \textbf{X-ray contraband decomposition}, \textbf{Transparent glassware decomposition}, \textbf{Semi-transparent cell decomposition}, \textbf{Semi-transparent watermark removal}, \textbf{Translucent occlusion removal}, and \textbf{Translucent flare removal}. Each sub-dataset contains around 5000-10000 training images and 300-500 test images. Each dataset simulates the visual phenomena of transparent and semi-transparent objects in the real world, such as occlusion, feature overlap, glare reflection, etc., by performing task-specific mixing of a foreground RGBA image with an alpha channel and a natural or structured RGB background. The image composition process follows the general formulation: $I = \mathcal{A}_\alpha \oplus_{\text{Blend}} \mathcal{B}$, where ${\mathcal{A}}_\alpha$ denotes the foreground image with an associated alpha transparency map, $\mathcal{B}$ is the background image, and ${\oplus}_{\text{Blend}}$ is a task-specific blending operator (e.g. alpha blending or screen mode). 

\textbf{X-ray Contraband Decomposition sub-dataset.} To emulate the physical property of X-ray in synthetic data, we formulate its process as: $I=(1-\alpha) \cdot \mathcal{B}+\alpha \cdot\left(\frac{\mathcal{A}_\alpha}{255} \cdot \frac{\mathcal{B}}{255} \cdot 255\right)$, where $\alpha$ controls foreground transparency. This nonlinear blending mimics the cumulative darkening seen in overlapping X-ray materials.

\textbf{Transparent Glassware Decomposition sub-dataset.} To simulate transparent and refractive materials like glassware, we formulate its process as: $I=\mathcal{A}_\alpha(x, y) \cdot \mathcal{B}(x, y)$, when the alpha channel of $\mathcal{B}(x, y)$ is lower than 0.5 and $I=1-(1-\mathcal{A}_\alpha(x, y)) \cdot(1-\mathcal{B})(x, y)$, when $\mathcal{B}(x, y)$ is higher than 0.5 during foreground-background composition. Unlike X-rays, such a condition should reflect real-world optical properties of transparent glass, allowing highlights to be intensified while preserving subtle background attenuation, thus producing a visually realistic rendering of refractive materials.

\textbf{Semi-transparent Watermark Removal sub-dataset.} 
To simulate real-world visual effects of semi-transparent overlays such as watermarks, logos, or textual elements, we formulate the compositing process as: $I=(1-\alpha(x, y)) \cdot \mathcal{A}_\alpha(x, y)+\alpha(x, y) \cdot \mathcal{B}(x, y)$. A linear alpha blending approach effectively reflects the smooth visual interference watermarks imposed on natural scenes while preserving contextual consistency for downstream layer decomposition.

\textbf{Semi-transparent Cell Decomposition sub-dataset.} For microscopic imaging, we simulate overlapping semi-transparent cell structures using a simple additive model: $I=\mathcal{A}_\alpha(x, y)+\mathcal{B}(x, y)$. Unlike X-ray or glass-like materials that rely on light attenuation or refractive blending, this additive formulation captures the cumulative intensity build-up commonly observed in microscopy.

\textbf{Translucent Occlusion Removal sub-dataset.} To simulate natural environmental occlusions such as fogged or rainy windows, this dataset overlays translucent layers onto outdoor or indoor background scenes. In contrast to object-level foreground overlays, this process is full-image, diffuse blending, where translucent elements span across the entire scene without distinct semantic boundaries. 

\textbf{Translucent Flare Removal sub-dataset.} The foreground lens glare or light reflection images are also synthesized with natural background images by using full-image diffuse blending. Overall, the translucent occlusion removal and the translucent glare removal datasets can be used as a benchmark for assessing generalizable, mask-free layer separation under ambient environmental interference.

Collectively, these datasets provide a comprehensive benchmark for evaluating models in realistic scenes. More details of our proposed AlphaBlend dataset can be found in the appendix.
\vspace{-0.1cm}
\subsection{Problem Formulation and Overview}
\begin{figure}[!t]
    \centering
    \includegraphics[width=0.9\textwidth]{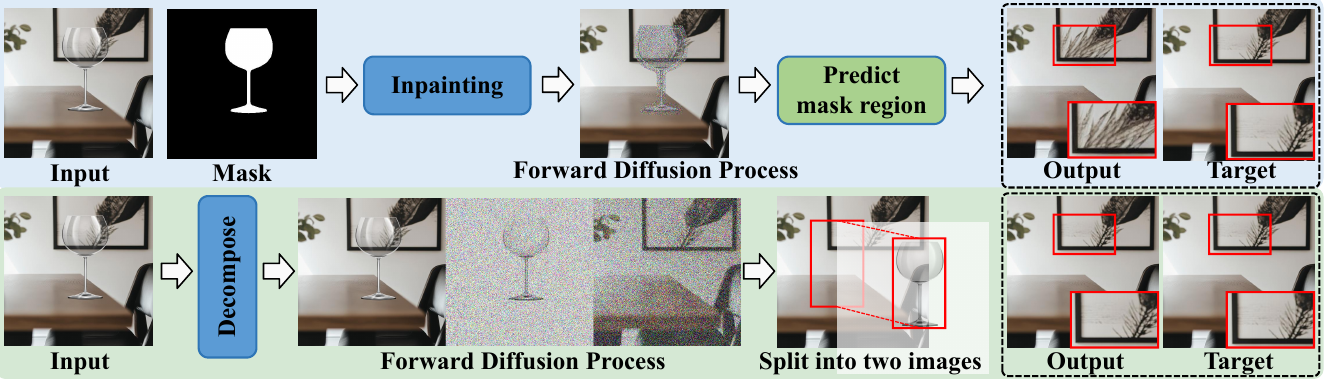}
    \caption{The comparison of conventional inpainting methods with our proposed DiffDecompose. The conventional inpainting approach (blue background) relies on predefined object masks and predicts missing regions, often causing semantic errors in transparent scenes. In contrast, DiffDecompose (green background) conditions on the full composited image and jointly predicts foreground and background via layer-level decomposition. This formulation removes the need for explicit masks and enables more accurate separation in the presence of transparency and complex blending. }
    \label{fig2}
    \vspace{-0.3cm}
\end{figure}

\noindent \textbf{Problem Formulation.} 
As shown in blue background of Figure~\ref{fig2}, to obtain the background from the observed image \( \mathbf{z} \in \mathbb{R}^{H \times W \times 3} \), the current method \cite{yu2023inpaint, sdxl, ekin2024clip} regards this problem as an image edition task or a common mask-based layer-wise decomposition task, which can be written as: 
\begin{equation}\label{equ2}
p_{\theta}(\mathbf{y} \mid \mathbf{z}, \mathbf{m})=\int p_{\theta}(\mathbf{y} \mid \mathbf{z}, \mathbf{m}) p_{\theta}\left(\mathbf{z}_{0} \mid \mathbf{z_t}, \mathbf{m}\right) d \mathbf{z}_{0},
\end{equation}
where $m$ represents the region that should be foreground split from the background, and the whole process can be viewed as the prediction of region $m$. The goal is to Unfortunately, since the semi-transparent/transparent object will refract the background information, including translucent object perception (e.g., glass or plastic), biological imaging (e.g., semi-transparent cells), and X-ray security scans, the pixel-level values of occluded region are no longer the simple coverage or replace, it is more of a nonlinear blend of multiple layers, which is totally different from existing tasks. Directly deleting the occlusion pixel and utilizing the background information to predict a new pixel is unreasonable and will result in false pixel-level restoration.

Considering such a common condition, we define it as a novel task, namely Layer-Wise Decomposition of Alpha-Composited Images, where the problem is decomposing an observed image that results from the layered composition of a semi-transparent/transparent foreground and a background. Formally, let \( \mathbf{x} \in \mathbb{R}^{H \times W \times 4} \) denote the foreground RGBA image, and \( \mathbf{y} \in \mathbb{R}^{H \times W \times 3} \) denote the RGB background image. These are combined through a (possibly nonlinear) composition function \( \mathcal{G} \), resulting in an observed image \( \mathbf{z} \in \mathbb{R}^{H \times W \times 3} \): $\mathbf{z} = \mathcal{G}(\mathbf{x}, \mathbf{y})$. The function \( \mathcal{G} \) may range from simple alpha blending to more complex operators such as additive, multiply, screen, or overlay composition. In practice, the exact formulation of \( \mathcal{G} \) may be unknown or variable across domains.

Thus, the decomposing background and foreground can be defined as a layer-wise decomposition process (the green background in Figure~\ref{fig2}). Given the observed image \( \mathbf{z} \), our goal is to recover a plausible pair of images \( (\mathbf{x}, \mathbf{y}) \) such that the composition function \( \mathcal{G}(\mathbf{x}, \mathbf{y}) \approx \mathbf{z} \). This inverse problem is highly ill-posed, particularly when \( \mathcal{G} \) is nonlinear and information is entangled across layers. To address this challenge, we propose to train a conditional diffusion model that learns the joint posterior distribution over foreground and background conditioned on the composed image, which can be defined as:
\begin{equation}
p_{\theta}(\mathbf{x}, \mathbf{y} \mid \mathbf{z},  \tau)=\int p_{\theta}(\mathbf{x}, \mathbf{y} \mid \mathbf{z}, \tau) p_{\theta}\left(\mathbf{z}_{0} \mid \mathbf{z}, \tau\right) d \mathbf{z}_{0}.
\end{equation}

During the training, we assume access to a dataset of triplets \( (\mathbf{x}, \mathbf{y}, \mathbf{z}) \), where \( \mathbf{z} = \mathcal{G}(\mathbf{x}, \mathbf{y}) \). The model is trained to sample decomposed pairs that are both semantically plausible and compositionally consistent under \( \mathcal{G} \).

\begin{figure}[!ht]
    \centering
    \includegraphics[width=0.9\textwidth]{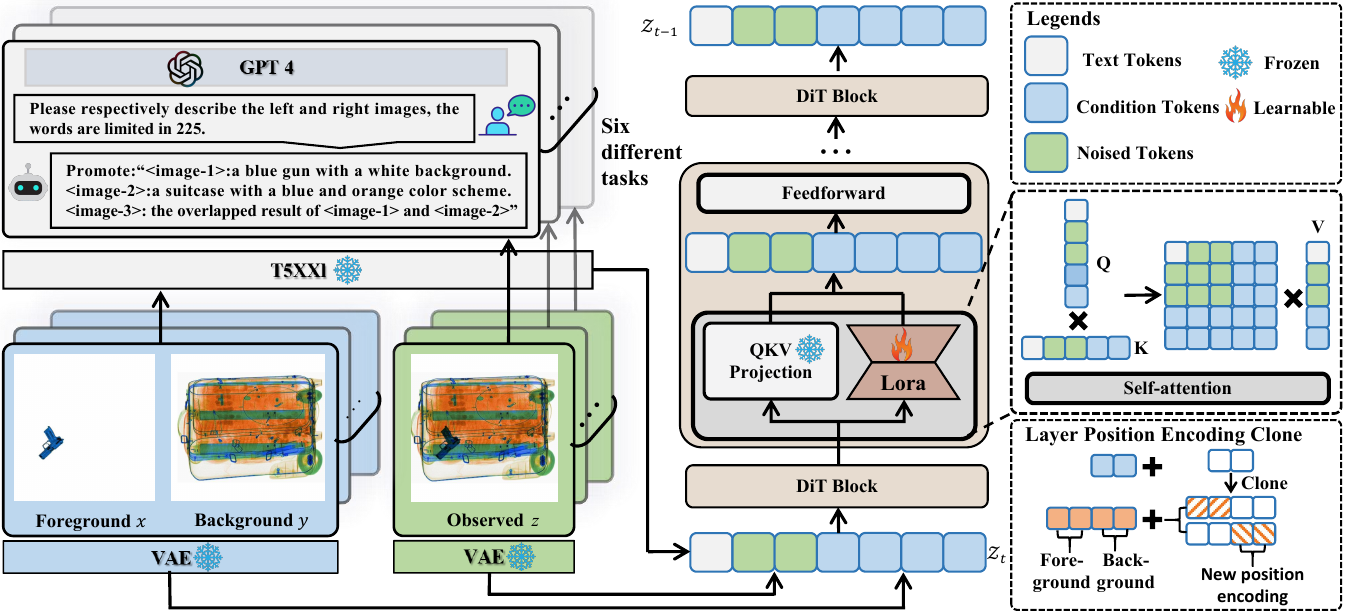}
    \caption{The DiffDecompose framework comprises two steps: (1) VAE encodes the source image into a condition token and concatenates it with a noisy latent token, controlling the generation of layer decomposition. (2) In-Context Decomposition constructs an image-conditioned base model to decompose multiple layers. Among them, the LPEC clones the position encoding to ensure the alignment between different layers, and MMAttention are introduced to process the multi-modulation features.}
    \label{fig3}
\end{figure}

\noindent \textbf{Framework Overview.} As shown in Figure~\ref{fig3}, the overview architecture of DiffDecompose can be divided into two steps: (1) Utilize the frozen VAE \cite{kingma2013auto} to respectively encode a semi-transparent/transparent foreground, a common RGB background, and a condition alpha-blend image, to obtain their latent spatial-level features. In the meantime, the prompts are fed into the frozen T5XXL \cite{raffel2020exploring} to extract the text-level features. (2) The process of decomposition is realized by In-context Decomposition (ICD) by concatenating the clean condition token and noise tokens, and utilizing the bidirectional attention to achieve conditional generation. During the process of ICD, we also devise a Layer Position Encoding Clone strategy to achieve consistency between the specific layer and the original layer information, and utilize the MMA for the interaction between different modalities.

\subsection{In-Context Decomposition}
Since the disentanglement of alpha-composited images is non-linear, directly learning the alpha maps is too difficult for decomposition. Thus, we develop an in-context decomposition to process layer decomposition as a context-aware spatial separation problem, where the model leverages both visual latent representations $f_x$, $f_y$, and $f_z$, and the textual latent representations $f_t$ to infer the spatial organization and semantic consistency of foreground and background components. To successfully realize this process, we develop the Layer Position Encoding Clone (LPEC) and introduce Multi-modality Attention (MMA) to achieve the process of in-context decomposition.

\vspace{-0.2cm}
\textbf{Layer Position Encoding Cloning.}
To enforce spatial coherence across compositional layers and enhance the disentanglement of foreground and background during generation, we propose the LPEC mechanism. Specifically, we compute the positional encoding ${\text{PE}}_z$ from the composited image $z$, and inject it into the token representations of the background image $y$ and composited observation $z$:
\begin{equation}
\tilde{c}_{y} = c_{y} + \mathrm{PE}z, \quad
\tilde{c}_{z} = c_{z} + \mathrm{PE}_z.
\end{equation}
This operation ensures that both the background and the composite tokens share a unified coordinate frame, i.e., $\forall(i, j), \quad \tilde{c}_{y}^{(i, j)}-\tilde{c}_{z}^{(i, j)}=c_{y}^{(i, j)}-c_{z}^{(i, j)}$ preserves relative positional structure during attention computation.
In contrast, the foreground tokens $c_{x}$ are excluded from this positional cloning, i.e.,$\tilde{c}_{x}=c_{x}$.
This separation avoids spatial entanglement between layers and prevents unwanted feature blending. In transparent or semi-transparent compositions, foreground and background often overlap spatially but differ in semantic meaning and transparency. Maintaining disjoint positional encodings: $\mathrm{PE}_{x} \perp \mathrm{PE}_{z}$, the model prevents spatial interference between foreground and background layers. By maintaining a distinct positional space for the foreground, the model is encouraged to preserve its semantic and spatial independence, thereby enhancing the fidelity of layer disentanglement in semi-transparent or fully transparent scenes.

\textbf{Multi-Modality Attention.} After applying LPEC, the latent tokens $c_x$, $c_y$, and $c_z$ are then concatenated along the sequence dimension to perform joint attention. The Multi-modal attention mechanisms \cite{esser2024scaling} are utilized to provide conditional information for the denoising of the alpha-composed image. Furthermore, by maintaining $y$ in a noise-free state, we ensure the retention of high-frequency textures and fine structural details from the original image, thereby preventing degradation during iterative denoising. This process can be written as:
\vspace{-0.1cm}
\begin{equation}
\mathrm{MMA}\left(\left[\tilde{c}_z; \tilde{c}_x; \tilde{c}_y; c_T\right]\right) =
\operatorname{softmax}\left(\frac{QK^{\top}}{\sqrt{d}}\right)V,
\end{equation}
where $Q, K, V$ are the query, key, and value projections of the concatenated sequence $[\tilde{c}_z; \tilde{c}_x; \tilde{c}_y; c_T]$, $d$ is the feature dimension, and $c_T$ represents the textual condition tokens extracted from a task-specific prompt. In particular, the description consists of the foreground image $x$, the background image $y$, and their composited observation $z$,  (e.g., “Three sub-images. <image-1>: a transparent glass, <image-2>: this is a dinning room with some chairs, <image-3>: The overlapped of <image-1> and <image-2>.), which guide the semantic understanding of the inter-layer relationships.

\section{Experiments}

\label{others}
\subsection{Experimental Settings}
\textbf{Implementation Details.} We initialize our model using the pre-trained parameters of the Flux.1~\cite{flux2024} architecture and fine-tune it across six representative subtasks. The model is trained using a LoRA \cite{esser2024scaling} rank of 128, batch size of 1, and a learning rate of ${10}^{-4}$ on a single H20 GPU for 30,000 steps. For the subtasks of translucent flare removal, translucent occlusion removal, semi-transparent watermark and removal, all input images are resized to 512×512. For the subtasks of transparent glassware decomposition, semi-transparent cell decomposition, and X-ray contraband decomposition, input images are resized to 512×1024, and the training is conducted for 50,000 steps under the same hyperparameter configuration unless otherwise specified.

\textbf{Baseline.} We compare our method with the state-of-the-art inpainting methods, which include the Inpainting Anything \cite{yu2023inpaint}, diffusion-based models SDXL-Inpaint \cite{sdxl}, Flux-control-Inpaint \cite{Controlnet}, PowerPaint \cite{zhuang2024task} and ClipAway \cite{ekin2024clip}. To ensure fair comparison and reduce inconsistencies or boundary artifacts, we apply mask dilation for five iterations using a 5×5 kernel prior to inpainting.

\textbf{Benchmark.} To provide a rigorous and comprehensive evaluation, we evaluate our DiffDecompose and baselines on two datasets: our proposed AlphaBlend dataset and the public LOGO dataset \cite{cun2021split}. For our proposed AlphaBlend dataset (described in Sec. \ref{sec3}), DiffDecompose infers its six scenarios, including: X-ray contraband decomposition, transparent glassware decomposition, semi-transparent cell decomposition, semi-transparent watermark removal, translucent occlusion removal, and translucent flare removal, with 300-500 test images. For LOGO evaluation, we utilize its three test datasets, LOGO-H, LOGO-L, and LOGO-G.
\vspace{-0.1cm}
\subsection{Main Results}

\begin{figure}[ht]
    \centering
    \includegraphics[width=0.94\textwidth]{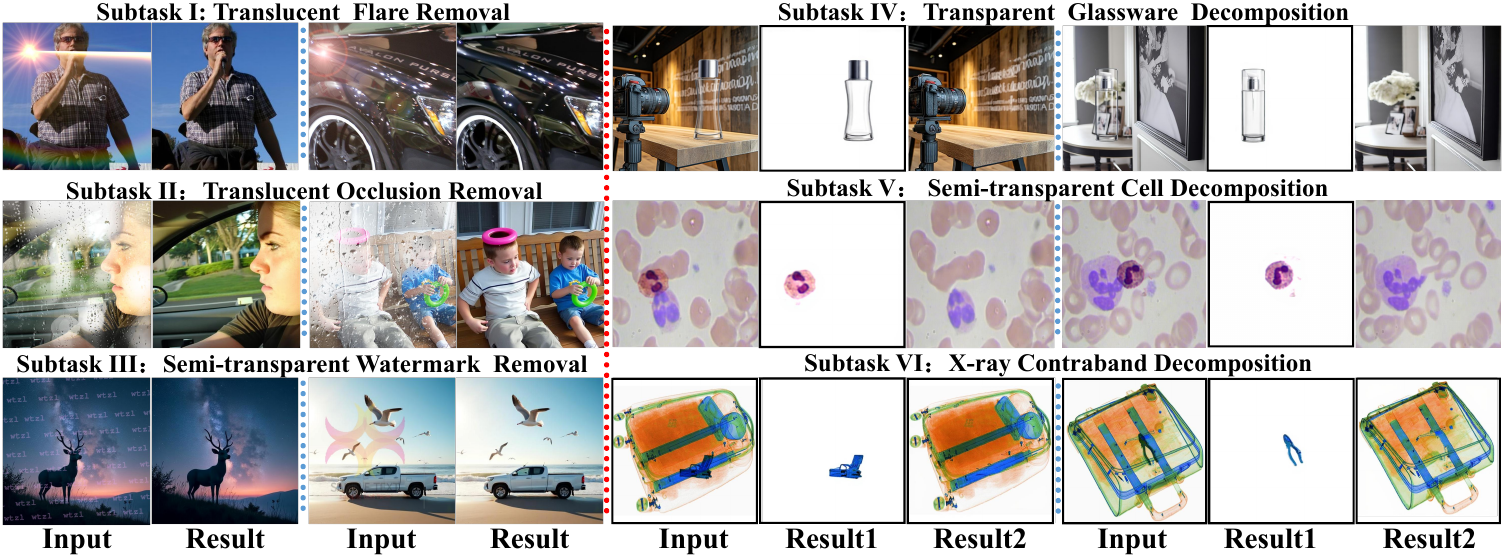}

    \caption{Our DiffDecompose shows impressive layer-level decomposition results of the image. It can solve the layer-level decomposition (i.e., Subtask II and Subtask III), the nonlinear alpha-blend layer removal and decomposition (i.e., Subtask I and Subtask VI), and the semi-transparent/transparent object-level decomposition (i.e., Subtask IV and Subtask VI), demonstrating its generalization and application in various scenarios.  }
    \label{fig4}
\end{figure}

Figure~\ref{fig4} showcases the performance of DiffDecompose across six presented subtasks. It can effectively disentangle complex compositional layers, ranging from window raindrops and lens flares to watermarks, glassware, X-ray contraband, and overlapping cells without reliance on explicit masks. It accurately reconstructs background content while preserving fine foreground structures under diverse transparency patterns and blending behaviors. These results demonstrate the method's strong generalization to real-world transparent and semi-transparent layer decomposition scenarios.

\begin{figure}[!ht]
    \centering
    \includegraphics[width=0.9\textwidth]{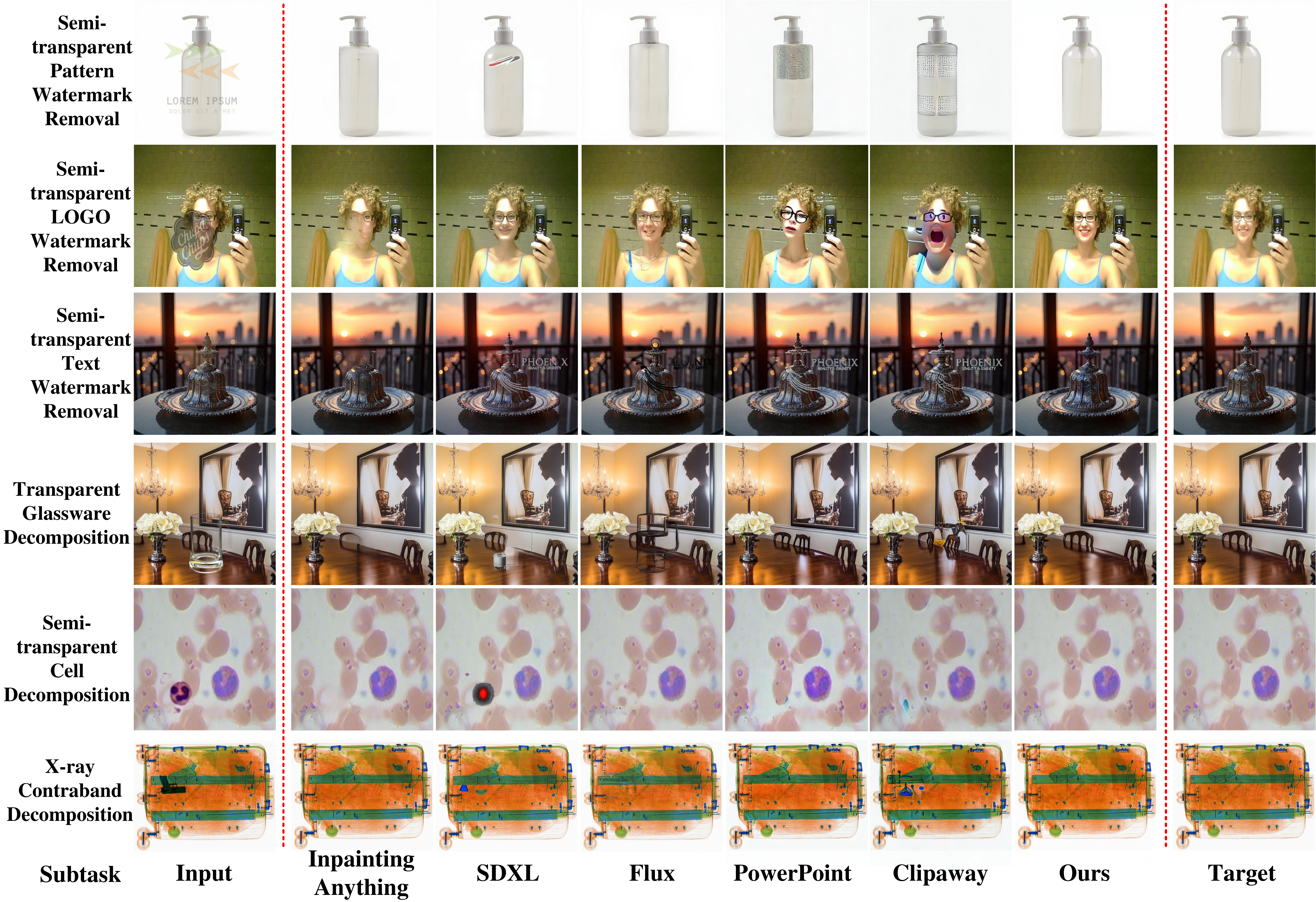}
    \vspace{-0.2cm}
    \caption{Qualitative comparisons between our method and other methods.
}
    \label{fig5}
\end{figure}

\vspace{-0.2cm}
\textbf{Qualitative Evaluation.} Figure~\ref{fig5} compares DiffDecompose with recent image inpainting-based SOTA methods across various semi-transparent layer removal tasks. While traditional inpainting approaches work for small, localized object-level occlusions, they struggle when foreground regions expand or exhibit transparency, often resulting in blurred textures or semantic inconsistencies. Methods like SDXL, ClipAway, and PowerPaint generate visually plausible but structurally inaccurate content, particularly in complex cases like transparent glass, X-ray contraband, or dense cell overlaps. In contrast, our DiffDecompose leverages layer-wise decomposition to disentangle foreground and background more accurately, preserving high-frequency textures and semantic continuity. Our method consistently outperforms other baseline methods in a range of challenging alpha-synthesized image scenarios, accurately recovering fine structural details, especially in watermark removal and transparent object decomposition. Furthermore, we found that these inpainting methods fail to remove the layer-level transparent occlusion, the details can be seen in Appendix.

\vspace{-0.2cm}
\textbf{Quantitative Evaluation.} Table~\ref{table1} quantitatively demonstrates the superiority of our DiffDecompose framework across both synthetic and public datasets, outperforming the second-best methods by an average margin of 36.3\% in RMSE, +1.2\% in SSIM, and 52.8\% in LPIPS. 

\begin{table}[!ht]
\vspace{-0.3cm}
  \caption{Comparison results with previous methods on LOGO and AlphaBlend dataset. }
  \label{table1}
  \centering
  \resizebox{\textwidth}{!}{
  \begin{tabular}{lcccccccccccc}
      \toprule
    Dataset  & \multicolumn{4}{c}{\textbf{LOGO-H}} & \multicolumn{4}{c}{\textbf{LOGO-L}} & \multicolumn{4}{c}{\textbf{LOGO-G}} \\
    \cmidrule(r){2-5} \cmidrule(r){6-9} \cmidrule(r){10-13}
     Models & RMSE$\downarrow$ & SSIM$\uparrow$ & LPIPS$\downarrow$ & FID$\downarrow$ & RMSE$\downarrow$ & SSIM$\uparrow$ & LPIPS$\downarrow$ & FID$\downarrow$ & RMSE$\downarrow$ & SSIM$\uparrow$ & LPIPS$\downarrow$ & FID$\downarrow$ \\
    \midrule
    PowerPoint \cite{zhuang2024task}& 10.0330 & 0.9726 & 0.0360 & 34.64 & 6.4480 & 0.9870 & 0.0212 & 30.49 & 8.1397 & 0.9811 & 0.0272 & 32.42 \\
    Flux-ControlNet \cite{Controlnet}& 11.7016 & 0.9640 & 0.0508 & 64.67 & 8.7095 & 0.9800 & 0.0351 & 61.22 & 14.5943 & 0.9670 & 0.0705 & 37.28 \\
    SDXL-inpainting \cite{sdxl} & 8.1936  & 0.9781 & 0.0302 & 35.24 & 4.9184 & 0.9899 & 0.0172 & 30.57 & 6.3711  & 0.9848 & 0.0221 & 32.37 \\
    ClipAaway \cite{ekin2024clip} & 11.3593 & 0.9690 & 0.0463 & 36.55 & 7.5353 & 0.9851 & 0.0275 & 32.40 & 9.3021  & 0.9785 & 0.0356 & 33.21 \\
    Inpaint Anything \cite{yu2023inpaint} & 4.8252  & 0.9849 & 0.0143 & 27.50 & 3.0461 & 0.9931 & 0.0072 & \textbf{23.71} & 3.8823  & 0.9898 & 0.0101 & 25.90 \\
    Ours   & \textbf{3.2321} & \textbf{0.9923} & \textbf{0.0092} & \textbf{27.25} & \textbf{2.1655} & \textbf{0.9962} & \textbf{0.0050} & 24.20 & \textbf{2.7940} & \textbf{0.9945} & \textbf{0.0063} & \textbf{24.78} \\
      \toprule
     Dataset  & \multicolumn{4}{c}{\textbf{Semi-transparent Watermark Removal}} & \multicolumn{4}{c}{\textbf{Transparent Glassware Decomposition}} & \multicolumn{4}{c}{\textbf{Semi-transparent Cell Decomposition}} \\
      \cmidrule(r){2-5} \cmidrule(r){6-9} \cmidrule(r){10-13}
       Models      & RMSE$\downarrow$ & SSIM$\uparrow$ & LPIPS$\downarrow$ & FID$\downarrow$ & RMSE$\downarrow$ & SSIM$\uparrow$ & LPIPS$\downarrow$ & FID$\downarrow$ & RMSE$\downarrow$ & SSIM$\uparrow$ & LPIPS$\downarrow$ & FID$\downarrow$ \\
      \midrule
      PowerPoint \cite{zhuang2024task} & 31.4009 & 0.8808 & 0.0571 & 59.70 & 10.6886 & 0.9727 & 0.0236 & 33.35 & 9.55 & 0.8504 & 0.1526 & 36.46 \\
      Flux-ControlNet \cite{Controlnet} & 35.9988 & 0.8430 & 0.1848 & 82.21 & 17.3372 & 0.9551 & 0.0451 & 68.41 & 12.08 & 0.9701 & 0.0407 & 89.47 \\
      SDXL-inpainting \cite{sdxl} & 21.7313 & 0.9163 & 0.0923 & 50.89 & 14.4375 & 0.9596 & 0.0421 & 73.36 & 9.70 & 0.9764 & 0.0274 & 79.18 \\
      ClipAaway \cite{ekin2024clip} & 35.3515 & 0.8641 & 0.1561 & 60.80 & 13.9394 & 0.9649 & 0.0333 & 47.64 & 7.58 & 0.9590 & 0.0180 & 49.79 \\
      Inpaint Anything \cite{yu2023inpaint} & 11.5273 & 0.9527 & 0.0484 & 34.58 & 8.2335 & \textbf{0.9775} & 0.0189 & 31.30 & 3.3028 & 0.9862 & 0.0093 & 36.42 \\
      Ours   & \textbf{2.9976} & \textbf{0.9894} & \textbf{0.0050} & \textbf{10.89} & \textbf{7.9938} & 0.9759 & \textbf{0.0134} & \textbf{25.55} & \textbf{2.4467} & \textbf{0.9901} & \textbf{0.0055} & \textbf{27.01} \\

      \toprule
       Dataset & \multicolumn{4}{c}{\textbf{X-ray Contraband Decomposition}} & \multicolumn{4}{c}{\textbf{Translucent Flare Removal}} & \multicolumn{4}{c}{\textbf{Translucent Occlusion Removal}} \\
      \cmidrule(r){2-5} \cmidrule(r){6-9} \cmidrule(r){10-13}
         Models    & RMSE$\downarrow$ & SSIM$\uparrow$ & LPIPS$\downarrow$ & FID$\downarrow$ & RMSE$\downarrow$ & SSIM$\uparrow$ & LPIPS$\downarrow$ & FID$\downarrow$ & RMSE$\downarrow$ & SSIM$\uparrow$ & LPIPS$\downarrow$ & FID$\downarrow$ \\
      \midrule
      PowerPoint \cite{zhuang2024task} & 7.1830 & 0.9826 & 0.0134 & 39.90 & 90.4829 & 0.1319 & 0.8150 & 404.70 & 89.1031 & 0.1336 & 0.8092 & 421.63 \\
      Flux-ControlNet \cite{Controlnet} & 7.6561 & 0.9808 & 0.1550 & 45.49 & 96.0870 & 0.2158 & 0.8016 & 253.40 & 87.8764 & 0.1991 & 0.8126 & 239.67 \\
      SDXL-inpainting \cite{sdxl} & 7.0060 & 0.9845 & 0.0125 & 39.38 & 63.1518 & 0.3539 & 0.7269 & 229.96 & 64.6848 & 0.3296 & 0.7351 & 209.20 \\
      ClipAaway \cite{ekin2024clip} & 9.9123 & 0.9789 & 0.0200 & 55.92 & 97.5711 & 0.2416 & 0.8362 & 255.26 & 96.2468 & 0.2344 & 0.8427 & 235.00 \\
      Inpaint Anything \cite{yu2023inpaint} & 4.8260 & 0.9856 & 0.0080 & 38.13 & 83.3769 & 0.3058 & 0.9040 & 462.58 & 83.2172 & 0.3090 & 0.9046 & 477.38 \\
      Ours   & \textbf{3.8903} & \textbf{0.9882} & \textbf{0.0049} & \textbf{23.73} & \textbf{16.4373} & \textbf{0.8189} & \textbf{0.0945} & \textbf{23.15} & \textbf{18.9561} & \textbf{0.7653} & \textbf{0.1445} & \textbf{43.116} \\
      \bottomrule
    \end{tabular}}
\end{table}
In complex real-world scenarios such as X-ray contraband removal and translucent window occlusions, our model achieves up to 4× lower RMSE and 3× better LPIPS than inpainting-based baselines, which struggle with color entanglement and global transparency. In particular, for challenging global occlusions like "Rain" and "Light", our method significantly surpasses all baselines, while others fail without explicit masks. Although our FID in LOGO-L is slightly higher than Inpaint Anything \cite{yu2023inpaint} (24.20 vs 23.71), which is attributed to the highly structured and mask-friendly layout of LOGO-L, providing strong mask priors. Importantly, our model is mask-free and shows consistently better structural accuracy across all LOGO datasets (e.g., RMSE ↓1.6, SSIM ↑0.002, LPIPS ↓0.011 over Inpaint Anything on LOGO-H). These results confirm that DiffDecompose not only generalizes across our six diverse datasets but also outperforms existing methods on public benchmarks, without requiring additional spatial priors.

\newpage

\subsection{Ablation study and Analysis}

\textbf{Effectiveness of LPEC.} As shown in Table~\ref{table2}, it can be seen that although LPEC is simple, it is crucial for different layer decomposition tasks. For the subtask of translucent occlusion removal at the layer-level, after studying the LPEC, it can improve 2.3/0.0242/0.0227 in terms of RMSE/SSIM/LPIPS. Apart from the layer-level decomposition, the object-level decomposition can also be viewed as the process of layer-level decomposition, and even more effectively. For the object-level semi-transparent Security, which is the most difficult for color distanglement, the LPEC can effectively separate the goods from their baggage. Additionally, the object-level transparent task of glass can improve the ICD results by 16.6106/0.2398/0.056 in terms of RMSE, SSIM, and LPIPS objective indicators. 

\begin{table}[!htbp]
\vspace{-0.2cm}
  \caption{Ablation of LPEC on different subtasks.}
  \label{table2}
  \centering
  \resizebox{0.9\textwidth}{!}{
  \begin{tabular}{lccccccccccccccccccccccccc}
      \toprule
    & \multicolumn{3}{c}{\textbf{X-ray Contraband}} & \multicolumn{3}{c}{\textbf{Semi-transparent glassware}} & \multicolumn{3}{c}{\textbf{Translucent occlusion}} & \multicolumn{3}{c}{\textbf{Translucent flare}}\\
        \cmidrule(r){2-13}
    Subtask & RMSE$\downarrow$ & SSIM$\uparrow$ & LPIPS$\downarrow$ & RMSE$\downarrow$ & SSIM$\uparrow$ & LPIPS$\downarrow$ & RMSE$\downarrow$ & SSIM$\uparrow$ & LPIPS$\downarrow$ & RMSE$\downarrow$ & SSIM$\uparrow$ & LPIPS$\downarrow$  \\

    \midrule
    w/o LPEC & 16.5379 & 0.7922 & 0.0489 & 24.6044 &0.7361 & 0.0694 & 21.2568& 0.7411 & 0.1672& 32.4708 & 0.7040 & 0.2352\\
    Ours  & 3.8903 & 0.9882 & 0.0049 & 7.9938 & 0.9759 & 0.013 &  18.9561 & 0.7653 & 0.1445& 16.4373 & 0.8189 & 0.0945\\
      \bottomrule
    \end{tabular}}
\end{table}

\begin{figure}[!ht]
\centering
\begin{minipage}[t]{0.5\textwidth}
    \centering
    \includegraphics[width=\textwidth]{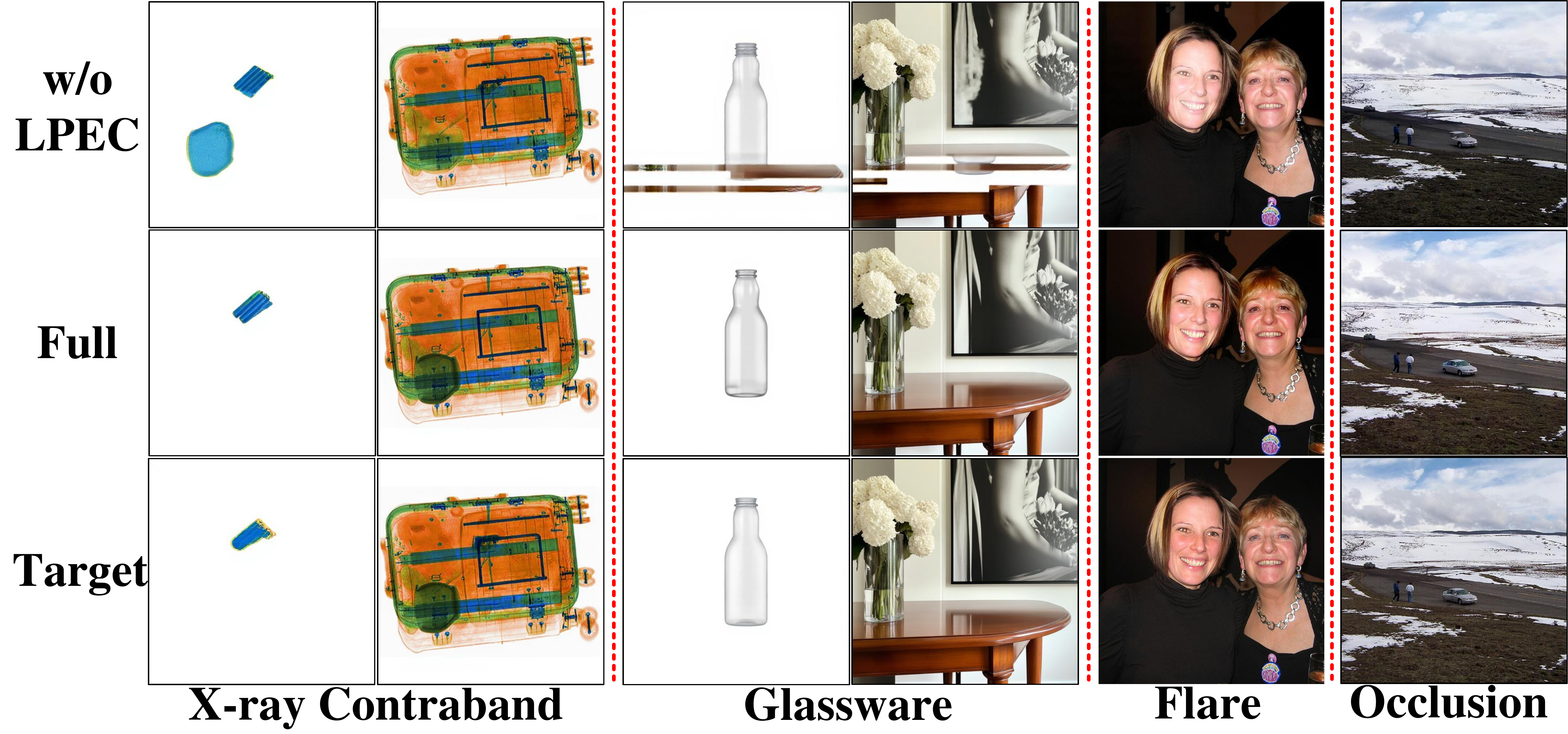}
        \vspace{-0.6cm}
    \caption{The visualization of LPEC ablation. Full settings ensure the information alignment of different layers, while removals degrade performance.}
    \label{fig6}
\end{minipage}%
\hfill
\begin{minipage}[t]{0.4\textwidth}
\centering
    \includegraphics[width=\textwidth]{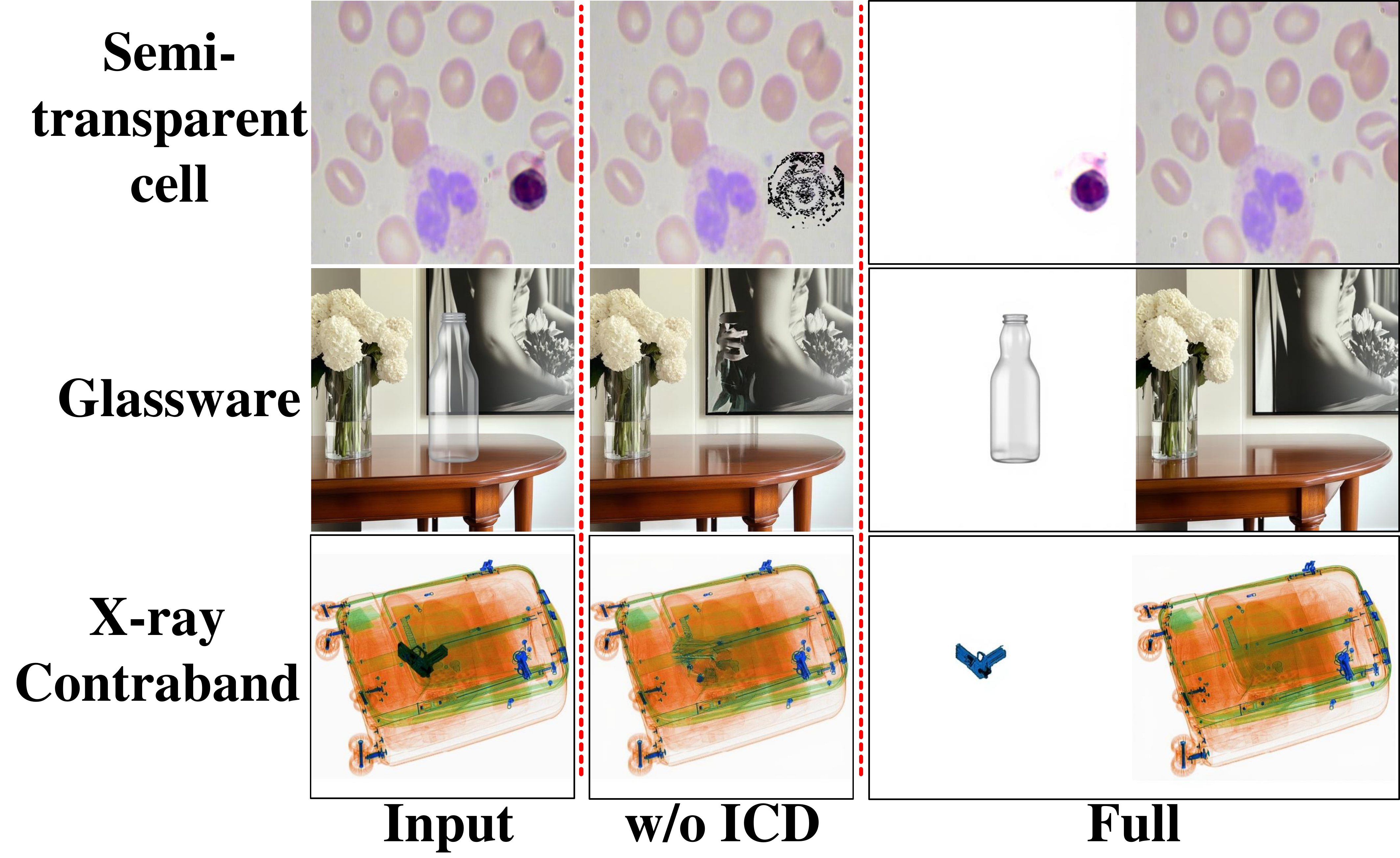}
        \vspace{-0.6cm}
    \caption{The visualization of ICD ablation. It shows that only the full setting enables effective decomposition. }
    \label{fig7}

\end{minipage}
\end{figure}

To further verify the effectiveness of LPEC, Figure~\ref{fig6} shows the results of DiffDecompose in four sub-tasks. As can be seen, when we remove LPEC, the layers of transparent glass and the most challenging subtask, X-ray contraband decomposition, will cause information distanglement. For the color of a person's face and the body of a car. More ablation studies are provided in the Appendix. 

\textbf{Effectiveness of ICD.}
The ICD can be regarded as the key to the task of decomposition. As shown in Figure~\ref{fig7}, when we totally remove the ICD, the framework fails to decompose the different layers and even results in a false prediction, which can be regarded as the failure of layer decomposition. In contrast, as we place the ICD on our baseline, the results can not only recover the accurate background but also output different layers. More ablation studies are provided in the Appendix. 

\vspace{-0.2cm}

\vspace{-0.5cm}
\section{Conclusion}
In this paper, we propose a novel layer decomposition task for the semi-transparent/transparent layer-wise decomposition of alpha-composited images. To support this new task, we propose a large-scale and high-quality AlphaBlend dataset that contains multiple layer-level transparent/semi-transparent foreground alpha-blend images. To cope with this complex task, we present a novel point that redefines the problem as learning a posterior over potential layer configurations conditioned on contextual cues. In the process of training, we propose a transformer diffusion-based framework, DiffDecompose, to learn the in-context decomposition to effectively predict single-layer or multi-layer results under given synthetic image conditions and achieve layer-by-layer decomposition. Furthermore, we also devise the LPEC for the process of ICD to effectively align the information between a certain layer and the original layer. Extensive experiments verify the effectiveness of semi-transparent/transparent layer-level decomposition task.
 
\newpage

% \small

{\small
\bibliographystyle{plain}
  \bibliography{main}
}

\newpage

\appendix
\textbf{Overview}
In the Appendix, we first present more details about our proposed dataset, AlphaBlend. Then, we introduce more information about the training and inference of DiffDecompose. Subsequently, we offer more competing results upon our proposed AlphaBlend dataset and a public dataset, LOGO, and also include additional quantitative experiments and complexity analyses. Finally, we present more visual results from generative results of DiffDecompose. The supplementary includes the following sections:
\begin{itemize}
    \item \textbf{\hyperref[dataset]{A.}}AlphaBlend dataset.
    \item \textbf{\hyperref[Loss]{B.}}Training Loss Function.
    \item \textbf{\hyperref[Inference]{C.}}Inference Algorithm.
    \item \textbf{\hyperref[Experiments]{D.}}More Comparison Experiments and Generative Results.   
    \item \textbf{\hyperref[User]{E.}}User Study and Broader Impacts.
    \item \textbf{\hyperref[Limitations]{F.}}Limitations.
\end{itemize}

\section{AlphaBlend dataset} \label{dataset}
\begin{figure}[!ht]
    \centering
    \includegraphics[width=\textwidth]{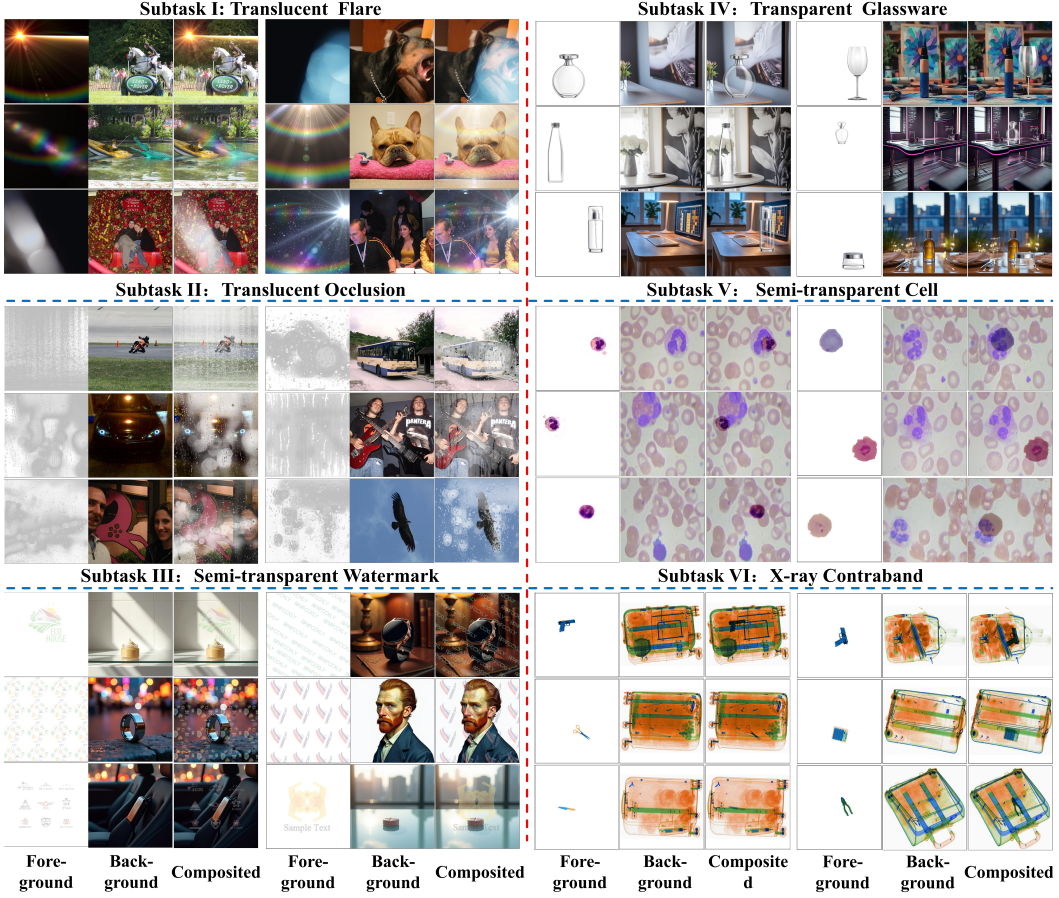}
    \caption{The presentation of the six subtasks' dataset. Each foreground has its respective properties.}
    \label{sup_fig1}
\end{figure}

As can be seen from Figure ~\ref{sup_fig1}, the properties of the AlphaBlend dataset are different from the existing datasets that are intended to generate a foreground layer object information that is completely opaque but has an alpha channel. The AlphaBlend datasets consider different semi-transparent conditions, from layer-level color distanglement to simple region-level transparent object occlusion. The task is no longer the same as the tasks of object removal \cite{yu2023inpaint, sdxl, ekin2024clip}, downstream image restoration \cite{lan2025exploiting, lin2025nighthaze}, or layer generation \cite{zhang2024transparent, yang2024generative}, and is also a common scenario in the real world. For tasks I and II, the rain, fog, light, and dirt can always occlude the whole background. Different from the common image dehazing and deraining, the pixel of the background has not been covered or fuzzy, but makes itself present semi-transparent. For task III, we further consider the more complex watermark condition in the real world, the current watermark dataset always presents lower opacity, and smaller occlusion, providing sufficient background information for inpainting methods to predict their background. Thus, we introduce bigger coverage with the size of 96$\sim$128, and the opacity below 0.25, which is more difficult to recognize. Tasks III, V, and VI, respectively, represent different pixel-level non-linear occlusions. These subtasks utilize part of the dataset \cite{zheng2018fast,alam2019machine,wei2020occluded,tan2025}.

\section{Training Loss Function} \label{Loss}
The loss function is formulated within the conditional flow matching \cite{flowmatch} framework as the joint expectation over timesteps, noisy latent variables, and noise samples:
\begin{equation}
L_{lce} = \mathbb{E}_{t, p_{t}\left(x_{t} \mid \epsilon_{x}\right), p_{t}\left(y_{t} \mid \epsilon_{y}\right), p\left(\epsilon_{x}\right), p\left(\epsilon_{y}\right)}\left\|v_{\Theta}\left(z, t, x, y, c_{T}\right)-u_{t}(z \mid \epsilon_{x},\epsilon_{y})\right\|^{2}
\end{equation}
where $v_{\Theta}\left(z, t, x, y\right)$ are the neural network’s predicted conditional velocity fields for the noisy foreground and background latent variables $x_t$ and $y_t$, respectively, conditioned on the timestep $t$, the observed composite image $z$, and conditioning information $\tau$. The true velocities $u_{t}(z \mid \epsilon_{x},\epsilon_{y})$, are derived from the diffusion process and noise samples $\epsilon_{x}$, $\epsilon_{y}$. This loss enables the model to learn an optimal probabilistic path from noise to the true decomposed images, thereby jointly inferring foreground and background layers under complex nonlinear composition.

\section{Inference Algorithm} \label{Inference}
Algorithm \ref{algorithm1} demonstrates the inference process of DiffDecompose. The composited image $z$ is regarded as the condition input of model $D(\cdot)$ without adding any noise, and foreground $x$ and background $y$ will be first randomly sampled by Gaussian distribution noise, and gradually generate their corresponding information guided by the prompt and condition image.
\begin{algorithm}

\caption{DiffDecompose Inference Algorithm}
\label{algorithm1}
\begin{algorithmic}[1]
\STATE \textbf{Input:} Composited Image $z$, Trained DiffDecompose model $D(\cdot)$ at a sampling timestep $t$
\STATE \textbf{Output:} foreground image $x$ and background image $y$
\FOR{epoch = 1 to $t$}
    \STATE $x_t \sim \mathcal{N}(0,\,I)$
    \STATE $y_t \sim \mathcal{N}(0,\,I)$
    \STATE $\hat{v}_x, \hat{v}_y = v_\theta({x}_t, {y}_t, t, z, \tau)$
    \STATE $    \sigma_t^2 = \eta \sqrt{\frac{1 - \bar{\alpha}_{t-1}}{1 - \bar{\alpha}_t}}
    \left(1 - \frac{\bar{\alpha}_t}{\bar{\alpha}_{t-1}} \right)$
    \STATE $    \mathbf{x}_{t-1} = \sqrt{\bar{\alpha}_{t-1}} \left(
    \frac{\mathbf{x}_t - \sqrt{1 - \bar{\alpha}_t} \cdot \hat{v}_x}{\sqrt{\bar{\alpha}_t}} \right)
    + \sqrt{1 - \bar{\alpha}_{t-1} - \sigma_t^2} \cdot \hat{v}_x + \sigma_t z_t $
    \STATE
    $\mathbf{y}_{t-1} = \sqrt{\bar{\alpha}_{t-1}} \left(
    \frac{\mathbf{y}_t - \sqrt{1 - \bar{\alpha}_t} \cdot \hat{v}_y}{\sqrt{\bar{\alpha}_t}} \right)
    + \sqrt{1 - \bar{\alpha}_{t-1} - \sigma_t^2} \cdot \hat{v}_y + \sigma_t z_t $

    \STATE $I_{out} \leftarrow D(Z_T, t)$
\ENDFOR
\end{algorithmic}
\end{algorithm}

\section{More Comparison Experiments and Generative Results} \label{Experiments}

\subsection{Evaluation Metric.} We use the Frechet Inception Distance (FID) \cite{heusel2017gans} to assess the photorealism of the generated images by comparing the source image distribution with the inpainted image distributions. To measure the accuracy of correct object removal, we introduce root mean squared error (RMSE)\cite{hodson2022root}, structural similarity index measurement (SSIM)\cite{wang2002universal}, and Learned Perceptual Image Patch Similarity (LPIPS) \cite{snell2017learning}.

\subsection{More Comparisons}
We present a comprehensive comparison of our DiffDecompose and other methods on a series of tasks within our newly established benchmark AlphaBlend and publicly watermark dataset LOGO. We present two competing results in each subtask. 

\begin{figure}[!ht]
    \centering
    \includegraphics[width=\textwidth]{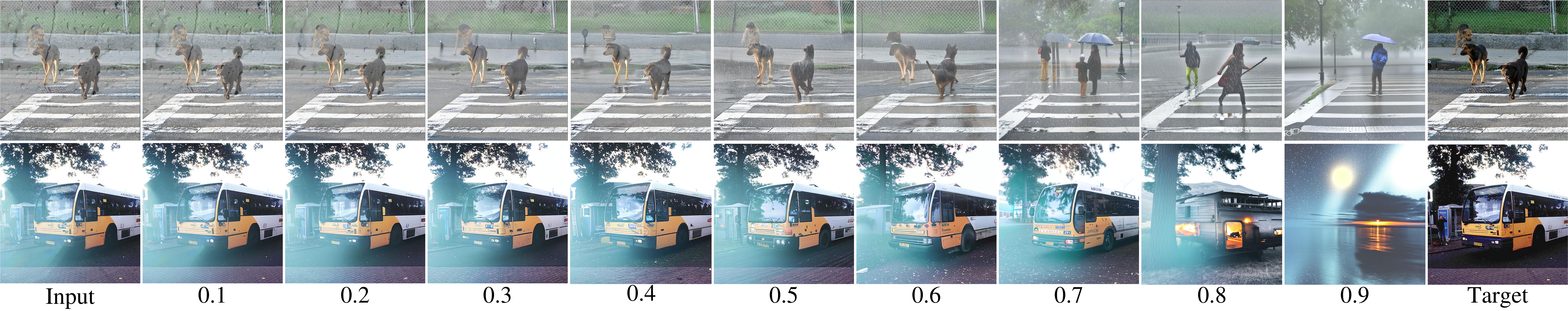}
    \caption{The SDXL-inpainting results under different strengths. We input the prompt like "Remove the light and make the scenario darker" and "Remove the foreground rain and fog", it is difficult for the model to process the layer-level edition.}
    \label{sup_fig8}
\end{figure}

For our proposed AlphaBlend, since the current inpainting methods~\cite{yu2023inpaint, sdxl, ekin2024clip} mainly rely on the mask-based method to achieve image inpainting, the task of occlusion and light removal is not reasonable for them to edit the whole image as shown in Figure~\ref{sup_fig8}, no matter how we tune the strengthen and change the prompt, it is still difficult to remove the rain and light. Thus, we only compare the results of the cell, glassware, and watermark, which can also be regarded as the region-level task instead of the layer-level task for these competing methods. It can be seen that when facing the transparent scenarios, the current methods directly predict the mask region with the background or prompt instead of fully utilizing the information of its transparent region. Additionally, when the occlusion region is pixel-additive, the competing method is prone to utilize the background to generate an unreasonable object to fix the removal cell. These results suggest that when the target object is no longer a simple pixel-level coverage, introducing the region-based methods for image editing is difficult to recover accurately. In comparison, our proposed DiffDecompose can better remove the foreground layer and maintain the background layer successfully.
To further verify the effectiveness of DiffDecompose, we extend the way to the common task of watermark removal. As shown in the latest three rows, the publicly available LOGO-G, LOGO-L, and LOGO-H, represent progressively challenging benchmarks for watermark removal, distinguished primarily by the transparency and size attributes of their embedded watermarks, these three public datasets can be concluded as:

\textbf{LOGO-G.} This dataset consists 2,000 testing samples, characterized by gray-scale watermarks. The watermark transparency in LOGO-Gray varies broadly from 35\% to 85\%, encompassing both relatively faint and highly opaque watermarks. The wide transparency range introduces substantial variability, simulating real-world scenarios where watermarks may range from subtle overlays to dominant visual artifacts. This dataset thus serves as a comprehensive testbed for evaluating models’ robustness across diverse watermark visibility levels.

\begin{figure}[!ht]
    \centering
    \includegraphics[width=\textwidth]{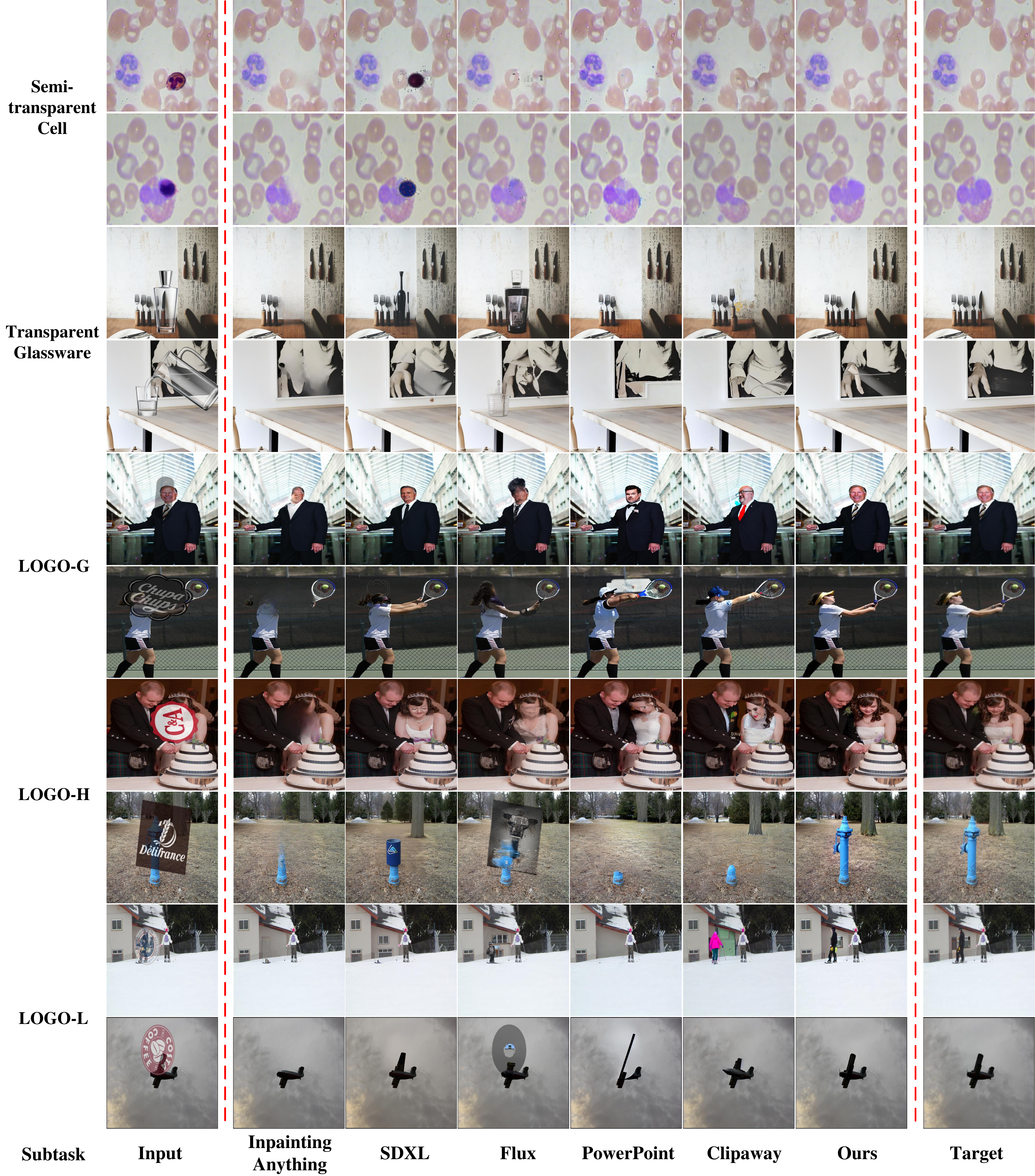}
    \caption{Qualitative results of our and compared methods on the proposed AlphaBlend dataset and the publicly LOGO dataset. Inpainting models often replace the object by predicting new pixel information instead of removing it, which fails to generate a realistic background. Our method is the only one that effectively removes objects and fills the regions in a accurate manner.}

    \label{sup_fig4}
\end{figure}

\textbf{LOGO-L.} LOGO-L restricts the watermark transparency to a narrower interval of 35\% to 60\%. Furthermore, the watermark size is constrained between 35\% to 60\% of the host image width. These constraints reflect moderate watermark prominence both in opacity and spatial extent, providing a balanced challenge for watermark removal techniques. The reduced transparency range relative to LOGO-Gray focuses the evaluation on semi-transparent watermarks, common in practical applications where watermarks are designed to be visible but non-intrusive.

\textbf{LOGO-H.} LOGO-H represents a more difficult subset derived from LOGO-L. It targets harder cases by increasing both watermark transparency and size, randomly selecting values from 60\% to 85\%. These higher transparency and larger size ranges imply that watermarks are more visually prominent and occlusive, significantly complicating the watermark removal task. Models evaluated on LOGO-H must contend with nearly opaque and spatially extensive watermarks, testing their ability to recover underlying image content under severe occlusion.

As shown in Figure \ref{sup_fig4}, our method outperforms existing approaches across challenging subsets, particularly LOGO-H and LOGO-L, which involve large occlusions and foregrounds with colors closely resembling the background. Compared methods often introduce artifacts or fail to fully remove watermarks under such conditions. In contrast, our approach achieves clearer and more coherent reconstructions by jointly inferring foreground and background through a layer-wise decomposition framework. This enables effective disentanglement of semi-transparent overlays without relying on explicit masks, making our method especially robust in complex, real-world scenarios involving transparency and subtle blending.

\subsection{More Generative Results}

\begin{figure}[!ht]
    \centering
    \includegraphics[width=\textwidth]{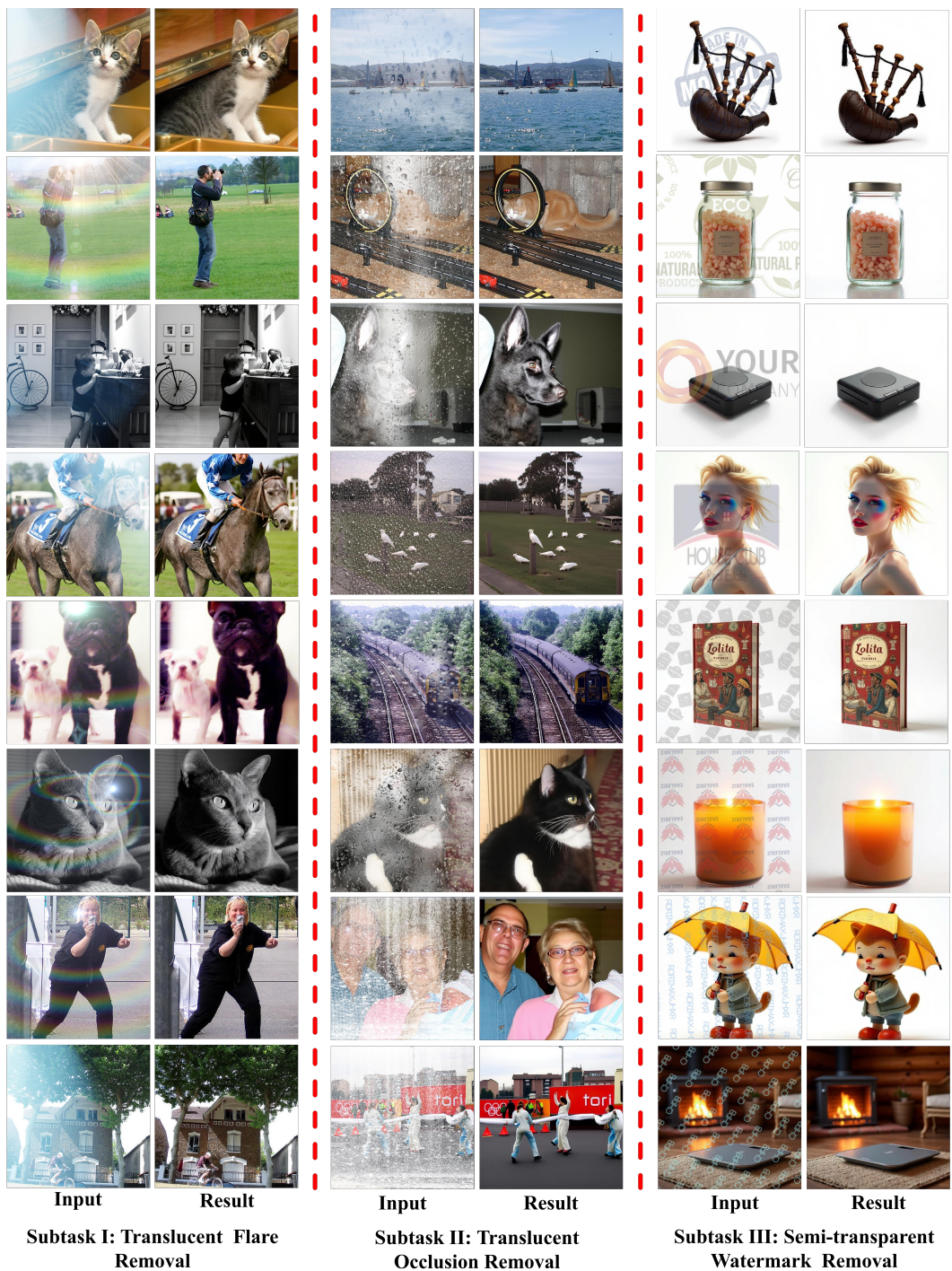}
    \caption{Additional Qualitative Results for Subtasks I–III. Extended visualizations demonstrating DiffDecompose’s capability in removing complex semi-transparent artifacts, including lens flares, translucent occlusions, and semi-transparent watermarks. The results illustrate faithful layer-wise decomposition, preserving fine details and semantic consistency across diverse real-world scenarios.}

    \label{sup_fig5}
\end{figure}

Figure \ref{sup_fig5} presents additional qualitative results for the first three subtasks of the AlphaBlend dataset: Subtask I (Translucent Flare Removal), Subtask II (Translucent Occlusion Removal), and Subtask III (Semi-transparent Watermark Removal). These extended visualizations further validate the capability of our DiffDecompose framework to disentangle complex semi-transparent and transparent layers in diverse real-world image compositions. DiffDecompose effectively removes complex lens flare while preserving scene structure and detail, separates translucent glass and reflection artifacts without disrupting background consistency, and accurately eliminates semi-transparent watermarks while maintaining underlying textures.

\begin{figure}[!ht]
    \centering
    \includegraphics[width=\textwidth]{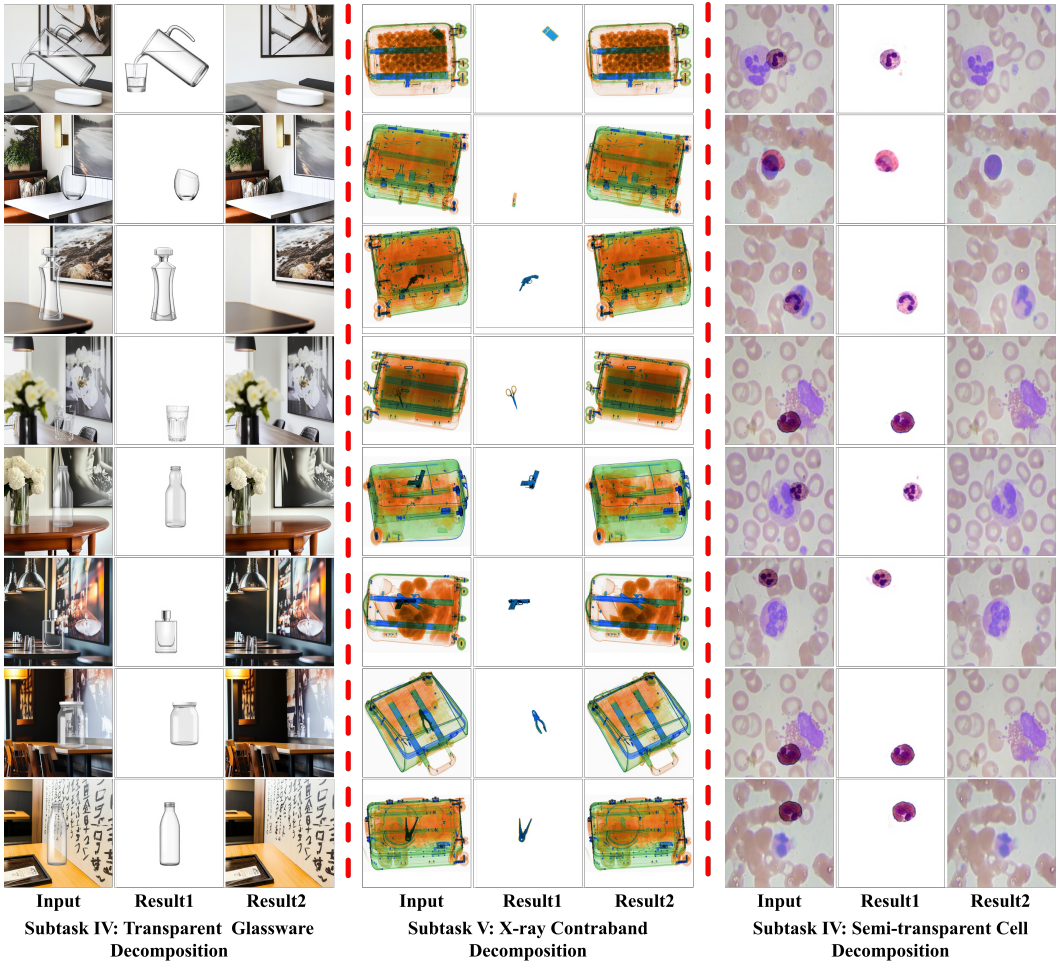}
    \caption{Additional Qualitative Results for Subtasks IV–VI. Extended results showcasing DiffDecompose’s effectiveness in transparent glassware decomposition, X-ray contraband separation, and semi-transparent cell decomposition. The visualizations highlight the framework’s ability to disentangle complex overlapping layers, preserving structural details and semantic fidelity in challenging transparency scenarios.}

    \label{sup_fig6}
\end{figure}

As shown in Figure \ref{sup_fig6}, we present more qualitative results of our proposed DiffDecompose framework on three key semi-transparent/transparent layer-wise decomposition subtasks from the AlphaBlend dataset: Subtask IV (Transparent Glassware Decomposition), Subtask V (Semi-transparent Cell Decomposition), and Subtask VI (X-ray Contraband Decomposition). These additional visualizations provide deeper insights into the robustness and generalization capabilities of our method in handling complex alpha-composited images characterized by nonlinear blending and ambiguous layer interactions. DiffDecompose isolates glass objects from varied backgrounds, accurately separates individual cells even in densely overlapping regions, and distinguishes hidden items from complex scan content while preserving contextual detail without relying on explicit masks.

As shown in Table \ref{sup_table1}, DiffDecompose achieves low RMSE and LPIPS scores alongside high SSIM across all three subtasks, indicating accurate structural and perceptual foreground reconstruction. These results confirm the method’s effectiveness in preserving fine-grained details and layer integrity, even under semi-transparent and complex blending conditions. By introducing a probabilistic layered decomposition framework, our model recovers visually coherent and semantically aligned foregrounds. This highlights the advantage of modeling compositional structures without requiring explicit pixel-level supervision.

\begin{table}[!ht]

  \caption{Quantitative Evaluation of Foreground Separation Quality. Performance metrics, including RMSE, SSIM, LPIPS, and FID, demonstrate DiffDecompose’s effectiveness in accurately extracting foreground layers across transparent glassware, X-ray contraband, and semi-transparent cell decomposition subtasks.}
  \label{sup_table1}
  \centering
  \resizebox{\textwidth}{!}{
  \begin{tabular}{lcccccccccccc}
    \toprule

    Subtask & \multicolumn{4}{c}{\textbf{Transparent Glassware Decomposition}} & \multicolumn{4}{c}{\textbf{X-ray Contraband Decomposition}} & \multicolumn{4}{c}{\textbf{Semi-transparent Cell Decomposition}} \\
    \cmidrule(r){2-5} \cmidrule(r){6-9} \cmidrule(r){10-13}
     Models & RMSE$\downarrow$ & SSIM$\uparrow$ & LPIPS$\downarrow$ & FID$\downarrow$ & RMSE$\downarrow$ & SSIM$\uparrow$ & LPIPS$\downarrow$ & FID$\downarrow$ & RMSE$\downarrow$ & SSIM$\uparrow$ & LPIPS$\downarrow$ & FID$\downarrow$ \\
    \midrule
    Ours & 0.7570 & 0.9993 & 0.0019 & 1.5463 & 0.7784 & 0.9994 & 0.0030 & 6.0607 & 0.2507 & 0.9997 &  0.0003 & 1.2532 \\
      \bottomrule
    \end{tabular}}
\end{table}

\clearpage

\subsection{User Study}

\begin{figure}[!ht]
    \centering
    \includegraphics[width=\textwidth]{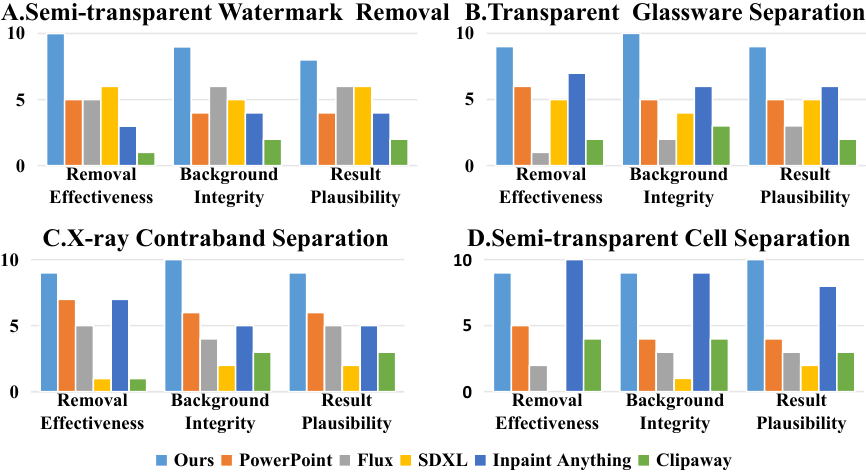}
    \caption{User study results. The voting results of DiffDecompose and the baseline method are compared on different subtasks, including removal effectiveness, background integrity, and result plausibility.}

    \label{sup_fig2}
\end{figure}

\begin{figure}[!ht]
    \centering
    \includegraphics[width=\textwidth]{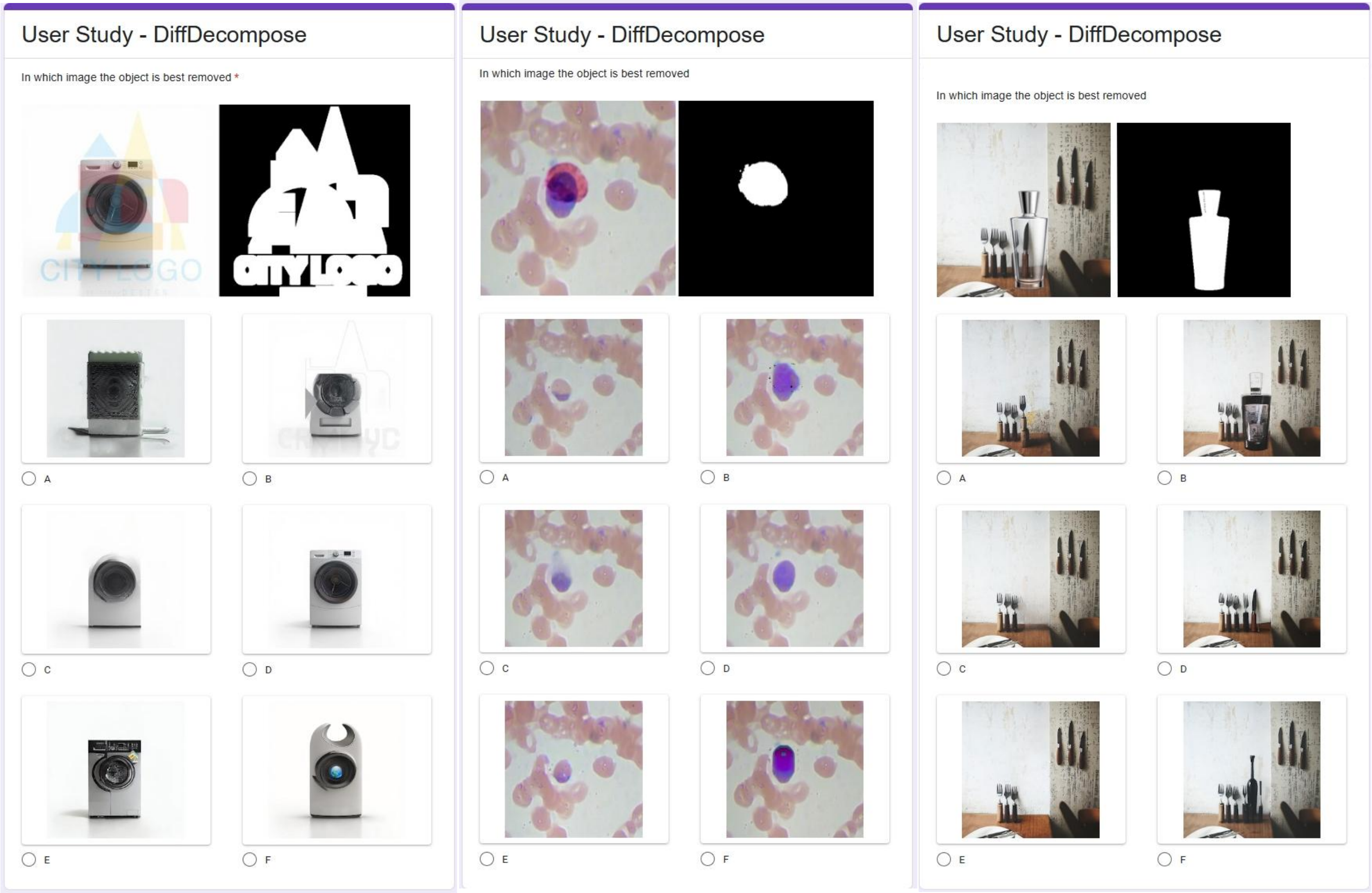}
    \caption{A user study voting interface was provided to participants. We present the performance of five competing methods on the task of watermark removal, cell separation, and glassware removal. The participants can click the alphabet to choose which method is the best and accord with their requirements. }

    \label{sup_fig3}
\end{figure}

To further evaluate the perceptual quality of our model compared to the baseline methods, we conducted a user study with 30 subjects using an online voting interface, the voting interface and voting results are shown in Figures \ref{sup_fig2} and \ref{sup_fig3}, respectively. For each visual scene, we presented the outputs of the six methods in a random order. Participants were asked to select the best result in terms of removal effectiveness, background integrity, and result plausibility for six representative subtasks. Our method received the highest number of votes in all six categories, demonstrating superiority over other competing baseline methods in terms of both realism and semantic coherence, without introducing unnecessary changes. 

\clearpage

\section{Limitations and Broader Impacts} \label{User}

\begin{figure}[!ht]
    \centering
    \includegraphics[width=\textwidth]{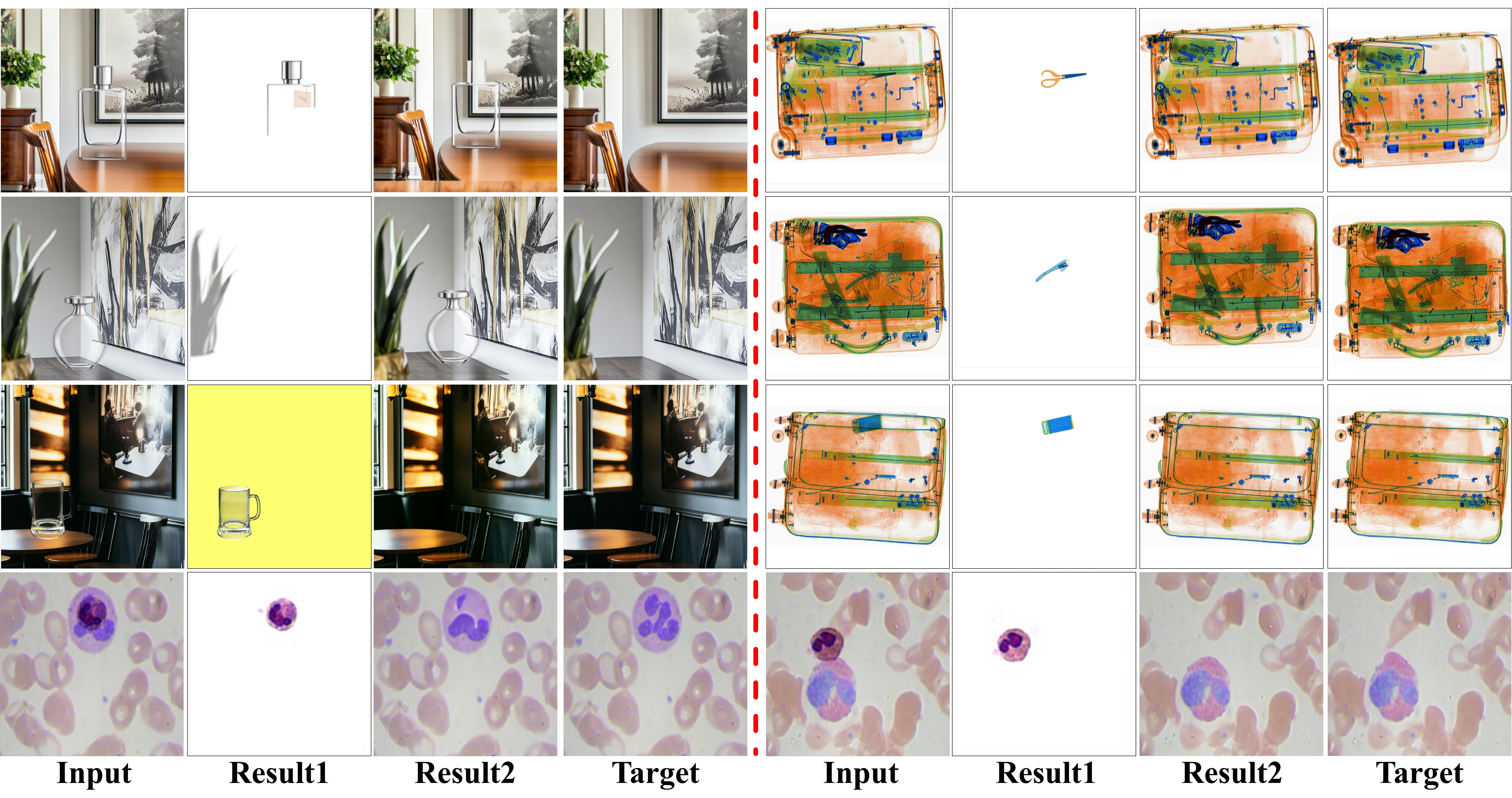}
    \caption{Results of failed decomposition by DiffDecompose.}

    \label{sup_fig7}
\end{figure}

\textbf{Limitations} \label{Limitations}
As shown in Figure \ref{sup_fig7}, although DiffDecompose performs well in various challenging semi-transparent and transparent layered decomposition tasks, it still has some limitations that deserve further study. Since DiffDecompose does not have an explicit supervision layer and fine-grained masks in the process of foreground and background separation, the pixels in the image repair area will have pixel drift that is imperceptible to the naked eye. That is, the overall separated foreground and background information have good visual contrast, but lack pixel-level position accuracy. Therefore, we will introduce Diffusion Guidance Masks in the future to improve the network's attention to image boundaries, thereby improving the accuracy of pixel restoration.

\textbf{Broader Impacts.}
This work may benefit applications in image editing, scientific visualization, and digital restoration by enabling accurate decomposition of semi-transparent and transparent layers. The public release of the AlphaBlend dataset and DiffDecompose code promotes reproducibility and supports broader research efforts. However, the ability to decompose image layers may also enable misuse in image manipulation or privacy invasion. We encourage responsible use and recommend the development of safeguards, particularly when extending this work to sensitive data. No personally identifiable or human subject data is involved in our experiments.

\end{document}